\documentclass[letterpaper]{article} 
\usepackage{aaai2026}  
\usepackage{times}  
\usepackage{helvet}  
\usepackage{courier}  
\usepackage[hyphens]{url}  
\usepackage{graphicx} 
\usepackage{natbib}  
\usepackage{caption} 
\usepackage{algorithm}
\usepackage{algorithmic}
\usepackage{amsmath} 
\usepackage{bm} 
\usepackage{amssymb}

\usepackage{multirow}
\usepackage{tabularray}
\usepackage{booktabs} 

\usepackage{graphicx}
\usepackage{subcaption}

\usepackage[table]{xcolor}  
\definecolor{mypink}{RGB}{255, 204, 204}     
\definecolor{mypale}{RGB}{255, 255, 204}     

\usepackage{newfloat}
\usepackage{listings}
\DeclareCaptionStyle{ruled}{labelfont=normalfont,labelsep=colon,strut=off} 
\floatstyle{ruled}
\newfloat{listing}{tb}{lst}{}
\floatname{listing}{Listing}
\title{MimicParts: Part-aware Style Injection for Speech-Driven 3D Motion Generation}

\author{
    Lianlian Liu\textsuperscript{1},
    Yongkang He\textsuperscript{2},
    Zhaojie Chu\textsuperscript{3},
    Xiaofen Xing\textsuperscript{4},
    Xiangmin Xu\textsuperscript{4}\thanks{Corresponding authors: texttt{xmxu@scut.edu.cn}}
}

\affiliations{
    \textsuperscript{\rm 1}South China University of Technology\\

}

\usepackage{bibentry}

\begin{document}


\twocolumn[{%
\renewcommand\twocolumn[1][]{#1}%
\maketitle
\begin{center}
    \centering
    \captionsetup{type=figure}
    \includegraphics[width=.99\textwidth]{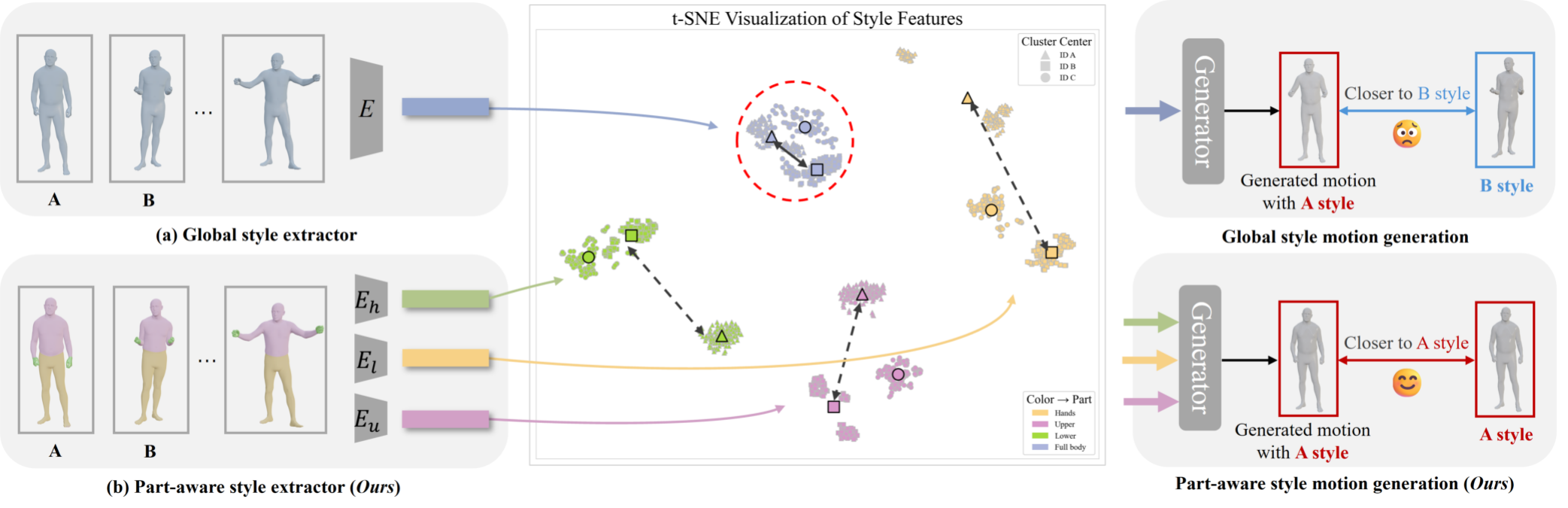}
    \captionof{figure}{Speech-driven stylized 3D motion generation aims to capture individual style differences and generate stylized motions synchronized with speech.
    Prior works often employ a global style extractor (a) that derives a single full-body style embedding from an motion sequence during training. This approach tends to capture only the overall motion style, overlooking regional style differences, which leads to reduced diversity and accuracy in generation. In contrast, our part-aware style extractor (b) focuses on learning finer-grained representations, enabling better capture and expression of regional style differences.
}
    \label{fig1}
\end{center}%
}]

\begin{abstract}
Generating stylized 3D human motion from speech signals presents substantial challenges, primarily due to the intricate and fine-grained relationships among speech signals, individual styles, and the corresponding body movements. Current style encoding approaches either oversimplify stylistic diversity or ignore regional motion style differences (e.g., upper vs. lower body), limiting motion realism. Additionally, motion style should dynamically adapt to changes in speech rhythm and emotion, but existing methods often overlook this.
To address these issues, we propose MimicParts, a novel framework designed to enhance stylized motion generation based on part-aware style injection and part-aware denoising network. 
It divides the body into different regions to encode localized motion styles, enabling the model to capture fine-grained regional differences.
Furthermore, our part-aware attention block allows rhythm and emotion cues to guide each body region precisely, ensuring that the generated motion aligns with variations in speech rhythm and emotional state.
Experimental results show that our method outperforming existing methods showcasing naturalness and expressive 3D human motion sequences. 


\end{abstract}


\section{Introduction}

Speech-driven 3D human motion generation is critical applications in VR, film and embodied AI.
As a core medium of non-verbal communication, expressive body gestures enhance speech communication effectiveness by conveying additional information such as thoughts, emotions, and intentions. A fundamental challenge in this domain is enabling virtual humans to transform emotion, rhythm, and personalized motion style into lifelike and expressive movements that authentically simulate human behavior.

Recent methods~\cite{yoon,MixStAGE,liu2022beat,yang2023diffusestylegesture,ghorbani2023zeroeggs,ao2023gesturediffuclip} have primarily achieved this goal through discrete style labels or learned style embeddings. However, these approaches are limited in capturing fine-grained stylistic differences, making it difficult to generate realistic full-body motion with detailed expressive styles.
As shown in Figure~\ref{fig1}(a), existing approaches~\cite{liu2022beat,yang2023diffusestylegesture,ghorbani2023zeroeggs,ao2023gesturediffuclip} for speech-driven stylized 3D motion generation primarily rely on two style encoding strategies: discrete one-hot encoding and global style extractor for holistic feature extraction. Nevertheless, both approaches have significant limitations. The one-hot encoding method~\cite{yang2023diffusestylegesture} struggles to effectively capture the inherent complexity and continuity of motion styles, and its limited representational capacity severely constrains model diversity and generalization. On the other hand, global style extractor~\cite{ghorbani2023zeroeggs,ao2023gesturediffuclip} ignore the differences in motion styles of different body parts.

To show differences of motion styles in different body parts, we train a contrastive style encoder~\cite{sun2024diffposetalk} that learns localized motion style representations, and visualized its style representations in the latent space. Figure~\ref{fig1} presents a t-SNE projection of the extracted features in the latent space.
The t-SNE projection reveals that while global style features of individuals are close in the latent space, localized style features show greater separability. This indicates that global style extractor blends style cues from different body parts, often overlooking localized differences and reducing motion diversity and accuracy. In contrast, part-aware style extractor capture more consistent and localized stylistic patterns.
Inspired by Jang et al.~\cite{Motionpuzzle}, we perform style encoding separately for different body parts.

On the other hand, previous studies on speech-driven stylized motion generation have mainly focused on improving the synchronization between motion and speech, or enhancing semantic consistency. These approaches typically rely on extracting simple speech embeddings~\cite{MixStAGE,ghorbani2023zeroeggs,yang2023diffusestylegesture}, or leveraging transcribed text to provide semantic information~\cite{ao2023gesturediffuclip}.
However, they generally neglect the influence of rhythm and emotion in speech on motion style modulation. Although existing methods can generate motions that partially respond to speech variations, the lack of dynamic style modulation limits their ability to fully reflect the rhythm and emotional dynamics of speech, thus constraining the expressiveness and naturalness of the generated motions.
Inspired by EmoTalk~\cite{emotalk} and AMUSE~\cite{AMUSE}, which model multi-level audio information, we extract content features, rhythm features, and emotion features from speech. We further model the modulation mechanism of emotion and rhythm on motion style, aiming to achieve more natural and expressive speech-driven stylized motion generation.

In this study, we propose MimicParts, a speech-driven gesture generation system that leverages a part-aware style encoder for part-aware style injection. As illustrated in Figure~\ref{fig1}(b), our framework first partitions reference motions into key body regions (upper body, lower body, and hands), and constructs a part-aware style encoder to extract localized motion style representations. 
Building on this, we introduce a part-aware diffusion model that precisely guides rhythm and emotion conditions to each region through part-aware attention block. This design effectively captures the modulation effects of rhythm and emotion on motion style.

Experimental results demonstrate that our method  outperforming current state-of-the-art methods in terms of style fidelity, motion-speech alignment, and perceptual naturalness.
In summary, our contributions are as follows:
\begin{itemize}
    \item We design a part-aware style encoder that extracts localized motion style representations from different body regions, enabling fine-grained modeling of localized stylistic differences.
    \item We propose a part-aware diffusion model that ensures the generated motion aligns with variations in speech rhythm and emotional state, thereby enhancing its naturalness and expressiveness.
    \item By integrating part-aware style modeling with a part-aware diffusion model, we construct the end-to-end MimicParts framework, which generates 3D human motion sequences that exhibit full-body coordination, naturalness, and expressiveness.
\end{itemize}




\section{Related Work}
\subsection{Speech-Driven Motion Synthesis}
Speech-driven stylized motion generation aims to synthesize motions that reflect speaker-specific styles or expressive variations from speech. Early works such as MixStAGE \cite{ahuja2020style} achieve multi-speaker motion generation by learning speaker style embeddings. 
ZeroEGGS~\cite{ghorbani2023zeroeggs} extracts global style features and generates stylized motions via simple feature concatenation, but its modeling struggle to capture semantic, emotional, and rhythmic dynamics, and overlooks localized style differences.
DiffuseStyleGesture \cite{yang2023diffusestylegesture} uses a diffusion model with one-hot style encoding but struggles to effectively represent complex style features. GestureDiffuCLIP \cite{ao2023gesturediffuclip} utilizes CLIP for prompt-based style control, but it faces issues with semantic interference \cite{gao2024styleshot}. 
Overall, current methods are limited by insufficient modeling of fine-grained localized style differences and lack dynamic adaptation to speech emotion and rhythm.

Speech-driven stylized motion generation primarily emphasizes expressive, individualized motions, whereas general speech-driven motion generation targets realistic, coherent full-body motions synchronized with speech.

Early methods \cite{ginosar2019learning,yoon2020speech,ahuja2020style,qian2021speech} focused on upper-body gestures conditioned on speech, often neglecting facial expressions or full-body coordination.
Habibie et al. \cite{habibie2021learning} first introduced full-body motion generation, but lacked inter-part coordination. 
To overcome these issues, DisCo \cite{liu2022disco} and HA2G \cite{liu2022learning} introduced generative models that decompose and hierarchically fuse speech features.
Building on this progress, CaMN \cite{liu2022beat} incorporated text semantics with facial and emotional cues. EMAGE \cite{liu2024emage} utilizes a four-branch combination VQ-VAE to generate expressive motions with speech rhythm and content features. 
Later studies further explore multimodal control,including AMUSE \cite{AMUSE} for emotion disentanglement and Syntalker \cite{chen2024enabling} for semantic-guided generation that synchronizes with speech and is consistent with text descriptions.

Based on the advances in speech-driven motion generation, we observe that existing methods often overlook the dynamic influence of speech on motion style. Inspired by this, we extract localized style representations and model the dynamic modulation of emotion and rhythm on motion style.

\subsection{Motion Generation via Diffusion Model}
Human motion generation has increasingly focused on generative models. Among them, diffusion models \cite{ao2023gesturediffuclip,yang2023diffusestylegesture,dabral2023mofusion}, known for their strong expressiveness and quality, are emerging as promising alternatives to GANs \cite{ahn2018text2action,lin2018human,li2022ganimator} and VAEs\cite{guo2020action2motion,petrovich2022temos,guo2022tm2t,yi2023generating}, especially in speech-driven and multimodal generation tasks.

 Initially, methods \cite{zhang2024motiondiffuse,yang2023diffusestylegesture,kim2023flame} focused on applying diffusion models to generate motion in the raw motion space. Recent methods \cite{ao2023gesturediffuclip,chen2023executing,chen2024enabling,yao2023dance} have shifted towards applying diffusion models in the motion latent space. In recent years, latent diffusion models have been introduced into speech-driven 3D motion generation to improve efficiency and multimodal modeling capabilities.
Syntalker~\cite{chen2024enabling} introduces a speech-driven motion model that applies discrete diffusion in the latent space of a Residual VQ-VAE~\cite{guo2024momask}.
GestureLSM \cite{gesturelsm} introduces flow matching technology, which achieves a more efficient sampling process by explicitly modeling the latent velocity space. Inspired by these methods, we propose a part-aware motion latent diffusion model for MimicParts, which allows rhythm and emotion cues to precisely guide each body region, ensuring that the generated motion aligns with variations in speech rhythm and emotional state.

\begin{figure*}[htbp]
  \centering
  \includegraphics[width=\textwidth,height=1\textheight, keepaspectratio]{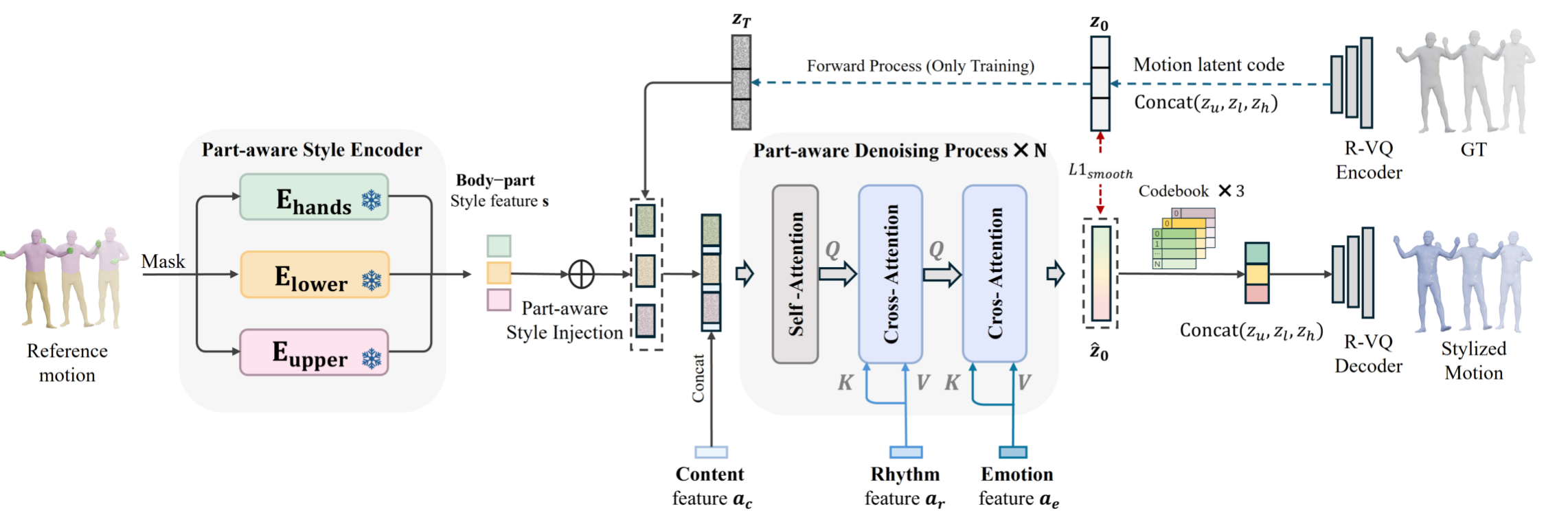}
  \caption{
    Overview of the MimicParts framework, which consists of a part-aware style encoder, a multi-level speech encoder, and a part-aware denoising module. 
    \textbf{Part-aware Style Encoder:} Given a reference motion and body-part masks, localized style representations for the upper body, lower body, and hands are extracted, enabling part-aware style injection. 
    \textbf{Part-aware Denoising Process:} Guided by speech features and localized style representations, the part-aware diffusion model progressively denoises \(z_T\). A part‑aware self attention block fuses concent features and local motion representations. Rhythm features and emotion features are then integrated via cross‐attention to guide each body region precisely.
    \textbf{R-VQ Decoder:} The final latent is quantized via body-part-specific codebooks and decoded into stylized 3D motion, achieving both full-body coherence and body-part-level stylistic diversity. 
  }
  \label{fig2}
\end{figure*}

\section{Method}

As shown in Figure~\ref{fig2}, we propose a speech-driven motion generation framework that produces stylized 3D motions guided by speech content, rhythm, emotion, and motion style. It operates in the latent space of an RVQ-VAE and leverages a part-aware style encoder and diffusion model to enable fine-grained, region-specific motion generation.

\subsection{Preliminary}

\textbf{3D Human Body Model.} 
We utilize the SMPL-X model \cite{pavlakos2019expressive} as our human body parameter model. SMPL-X is a widely recognized parametric human body model that represents the entire human shape, including detailed facial expressions, hand poses, and body posture. The SMPL-X model comprises $N=10,475$ vertices and $K=54 $joints, encompassing joints for the neck, jaw, eyeballs, and fingers.The SMPL-X model is defined by the function: 
\begin{equation}
M(\theta, \beta, \psi): \mathbb{R}^{|\theta| + |\beta| + |\psi|} \longrightarrow \mathbb{R}^{3N}
\label{eq:motion_decoder},
\end{equation}
where $\theta$ represents the pose parameters, $\beta$ denotes the shape parameters, and $\psi$ corresponds to the facial expression parameters.Given a set of parameters $\left( \theta, \beta, \psi \right)$,the function returns the coordinates of a human shape in a fixed order.

\noindent\textbf{Motion Representation.} 
Recent studies have shown that residual quantization (RQ) \cite{borsos2023audiolm}\cite{martinez2014stacked}significantly reduces the information loss caused by quantization in VQ-VAEs\cite{guo2024momask, zhang2023generating}, improving the quality of human motion reconstruction. Following \cite{chen2024enabling}, to reduce the coupling between different body parts, we divide the body into three regions: upper body, hands, and lower body, and pretrain separate RVQ-VAEs for each part.
Starting with the initial residual \( r^1 = \{ z_i \mid i = 1, 2, \dots, n \} \), the first quantization module produces the quantized output \( z^1 = Q(r^1) \) via nearest codebook lookup. The next residual is computed as \( r^2 = r^1 - z^1 \), which is then quantized by the second layer, and so on. Residual quantization (RQ) proceeds layer-by-layer as follows:

\begin{align}
z^v &= Q(r^v), \label{eq:quantization} \\
r^{v+1} &= r^v - z^v. \label{eq:residual}
\end{align}

Training uses reconstruction and per-layer latent losses:
\begin{align}
\mathcal{L}_{rvq} = \|\mathbf{m} - \hat{\mathbf{m}}\|_1 + \beta \sum_{v=1}^{V} \|\mathbf{r}^v - \text{sg}[\mathbf{z}^v]\|_2^2,
\end{align}
where $sg[.]$ denotes the stop-gradient operation, and $\beta$ a weighting factor for embedding constraint. 

\noindent\textbf{Speech Representation.} 
A large number of studies have shown that existing pre-trained speech models can effectively extract speech rhythm, content, and emotion features. Accordingly, we use BEATs \cite{chen2022beats}, Wav2Vec2 \cite{baevski2020wav2vec}, and Emotion2Vec \cite{ma2023emotion2vec} to extract rhythm, content, and emotion features, respectively. These features are jointly fed into our part-aware motion latent diffusion model to drive the generation of stylized motion.

\subsection{Part-aware Style Extractor}

As shown in Figure~\ref{fig2}, part-aware style encoder consists of three core modules, each corresponding to a different body part. Each part's style encoder is constructed based on the transformer encoder to extract style representations from the motion parameter sequence \(X= \left \{ x_i \mid i = 1,2,\ldots,n \right \}  \) of the corresponding part.

Due to the difficulty of quantifying motion styles using explicit labels or fixed templates, we adopt an implicit modeling approach based on contrastive learning to capture subtle stylistic differences. Our core assumption is that motion styles remain consistent within short, temporally adjacent segments performed by the same individual. Based on this assumption, we enhances the discriminative power and robustness of style representations by pulling together samples of the same style while pushing apart those of different styles.


Specifically, we employ the NT-Xent loss\cite{chen2020simple} to train the style encoder. Each training batch consists of $N_s$ samples, with each motion parameter sample having a length of $2T_s$. Each sample is split in half, resulting in $N_s$ positive pairs. Given one positive pair, the remaining $2(N_s-1)$ samples are treated as negative examples. We use the cosine similarity function, and the loss for a positive pair is defined as follows:
\begin{align}
\mathcal{L}_{i,j} = -\log \frac{\exp \left( \text{cos\_sim}(s_i, s_j) / \tau \right)}{\sum_{k=1}^{2N_s} \mathbf{1}_{[k \neq i]}\exp \left( \text{cos\_sim}(s_i, s_k) / \tau \right)},
\end{align}

where $ 1_{k \neq i}$ is an indicator function, $\tau$ denotes the temperature parameter. The overall loss is computed across all positive pairs for both$(i,j)$ and $(j,i)$.
Once pretrained, the part-aware style encoder serves as a style representations extractor in the latent diffusion model.

\subsection{Part-aware Motion Latent Diffusion Model}

As pointed out by previous studies \cite{zhong2023attt2m}, the jointed whole-body structure exhibits complex spatio-temporal relationships, calling for fine-grained modeling across different body parts.  In addition, existing methods overlook the fact that motion style should dynamically adapt to variations in speech rhythm and emotion, which constrains the naturalness of the generated motions.
Therefore, we propose Part-Aware Attention Block to explicitly model the unique dynamics of different body regions and enable fine-grained, region-specific modulation based on external conditions such as speech rhythm and emotion.


\begin{figure}[ht]
  \centering
   \includegraphics[width=0.5\textwidth]{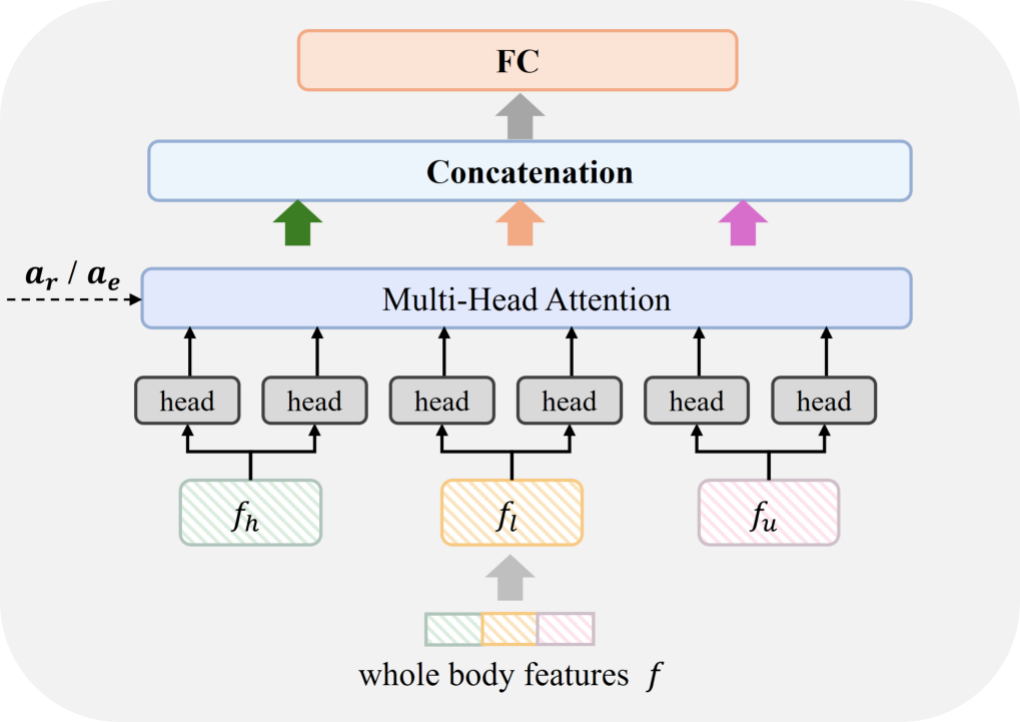}
  \caption{
Architecture of \textbf{Part-Aware Attention Block.} It allocates different attention heads to specific body regions, ensuring regional modeling, where \( f_u \), \( f_l \), and \( f_h \) represent features of the upper body, lower body, and hands, respectively. During cross-attention, rhythm ($a_r$) or emotion ($a_e$) features are used as the keys and values.
  }
  \label{fig:6}
\end{figure}

\noindent\textbf{Part-Aware Attention Block.}
Our diffusion model consisting of several part-aware attention blocks. 
As shown in Figure~\ref{fig:6}, the part-aware attention block allocates different attention heads to specific body regions, enabling fine-grained modeling of localized motion patterns. we split whole body into three different region, i.e. $f \to \{f_u, f_l, f_h\}$, where $ u $, $ l $, and $ h $ denote the upper body, lower body, and hands respectively. Each attention head focuses on a separate body part, and then the outputs of each head are fused through a fully connected layer:

\begin{equation}
f'=\mathrm{FC}\left( \operatorname{Concat}\left[ \mathrm{Att}_u(f_u),\ \mathrm{Att}_h(f_h),\ \mathrm{Att}_l(f_l) \right] \right),
\label{eq:part_attention_fusion}
\end{equation}
where the attention module \( \mathrm{Att}_i \) represents either a self-attention module \( \mathrm{Att}_i^s \) or a cross-attention module \( \mathrm{Att}_i^c \), depending on the conditioning modality, and is applied to the corresponding localized features \( f_i \).

\noindent\textbf{Denoising Network.}
We first generate motions that satisfy both semantic and stylistic requirements by leveraging content features \( a_c \) and localized style representations \( s_i \ (i \in \{u, l, h\}) \), integrated through a part-aware self-attention block. The fusion of content and localized style features is formulated as:
\begin{align}
z_i' = \operatorname{Concat}(z_i + s_i,\, a_c), \quad i \in \{u, l, h\},
\end{align}
where \( z_i \) denotes the localized latent motion representations.
Subsequently, as illustrated in Figure~\ref{fig2}, rhythm features \( a_r \) and emotion features \( a_e \) are progressively incorporated into different body regions via a part-aware cross-attention block, further enhancing the rhythmic consistency and emotional expressiveness of the generated motions.

Overall, our denoising network takes as input a set of multimodal conditions \((s, a_c, a_r, a_e)\), where \(s\) denotes the localized style representations, \(a_c\) denotes speech content features, \(a_r\)  denotes rhythm features, and \(a_e\) denotes emotion features:
\begin{align}
z_0 = D_\theta(z_t, t,s, a_c, a_r, a_e),
\end{align}
where \( z_0 \) denotes the latent representation encoded from the input motion sequence, and \( z_t \) denotes the noised latent at timestep \( t \in \{1, \ldots, T\} \). The loss function is defined as follows:
\begin{align}
\mathcal{L}_{net}=L1_{smooth}\ [z_0-D_\theta\ (z_t,t,s,a_c,a_r,a_e)].
\end{align}
By leveraging the part-aware attention block, the denoising network \( D_\theta \) effectively incorporates semantic, stylistic, rhythmic, and emotional cues across different body regions.

\noindent\textbf{Training Strategy.}
The training of our model follows a two-stage strategy. In the first stage, we pretrain the RVQ-VAE motion encoder and decoder, along with the part-aware style encoder. Given an input motion sequence \(X= \left \{ x_i \mid i = 1,2,\ldots,n \right \}  \), the pretrained encoder produces a latent representation $z_0 = E(X)$.  The part-aware style encoder, used to extract localized style representations \( s \), is kept frozen during diffusion model training.

In the second stage, we train the full diffusion model in the pretrained motion latent space.
The diffusion process gradually adds Gaussian noise to the latent motion features \( z_0 \), transforming them into a standard Gaussian distribution \( \mathcal{N}(0, \mathbf{I}) \) via a Markov forward process:
\begin{align}
q(z_t \mid z_{t-1}) = \mathcal{N}\left( \sqrt{\alpha_t} \, z_{t-1},\, (1 - \alpha_t)\, \mathbf{I} \right),
\end{align}
where \( \alpha_t \) controls the noise level. The reverse process learns to denoise \( z_t \) and progressively reconstruct \( z_0 \) via \( q(z_{t-1} \mid z_t) \).

\noindent\textbf{Sampling with Incremental Classifier-Free Guidance.}
During inference, the localized style representations $s$ is extracted from the reference motion, while the content features $a_c$, rhythm features $a_r$, and emotion features $a_e$ are derived from the target audio. Given these guidance conditions, we initialize the process with random noise $z_T$ and iteratively denoise it from timestep $T$ to $1$.  After the final step, the predicted latent $\hat{z}_0$ is decoded by the pretrained RVQ-VAE decoder to reconstruct the stylized motion sequence $\hat{X}= D(\hat{z}_0)$ .

To enhance controllability during inference, we further adopt classifier-free guidance with an incremental scheme, which has proven effective in multi-condition generation. The denoising function is computed as:
\begin{align}
&D_\theta(z_t,t,a_c,s,a_{re})  = \; D_\theta(z_t,t,\oslash,\oslash,\oslash) \nonumber \\
& + w_c \bigl(D_\theta(z_t,t,a_c,\oslash,\oslash) - D_\theta(z_t,t,\oslash,\oslash,\oslash) \bigr) \nonumber \\
& + w_{s} \bigl(D_\theta(z_t,t,a_c,s,\oslash) - D_\theta(z_t,t,a_c,\oslash,\oslash) \bigr) \nonumber \\
& + w_{re} \bigl(D_\theta(z_t,t,a_c,\oslash,a_{re}) - D_\theta(z_t,t,a_c,\oslash,\oslash) \bigr).
\end{align}
Here, \( w_c \) and \( w_s \) denote the guidance weights for content and style, respectively, while \( w_{re} \) controls both rhythm and emotion. To simplify notation, we define a unified representation for rhythm and emotion features as \( a_{re} \in \{a_r, a_e\} \). During training, conditional inputs are randomly dropped (i.e., set to \( \oslash \)) with a probability of \( p = 0.1 \) to jointly learn conditional and unconditional distributions. During inference, the degree of stylization in the generated motion can be flexibly controlled by adjusting the guidance weights.

\section{EXPERIMENTS}
\subsection{Experimental Setup}
\textbf{Dataset.} BEAT2, proposed in EMAGE~\cite{liu2024emage}, contains 76 hours of data from 30 speakers, each associated with eight emotional styles. We use the BEAT2-standard subset with an 85\%/7.5\%/7.5\% split for training, validation, and testing per identity. For the single-speaker model comparison, we select subject ID 2. For the multi-speaker model comparison, we use subjects with IDs from 1 to 10.

\noindent\textbf{Implementation Details.} We use the RVQ-VAE~\cite{chen2024enabling} as our auto-encoder, featuring residual blocks in both the encoder and decoder with a downscale factor of 4. Residual quantization utilizes 6 layers, each with a 512-dimensional codebook and a dropout ratio of 0.2.
For the local body style encoder, we employ a 4-layer Transformer with 4 attention heads, a feature dimension of $d_s = 512$, sequence length $T = 128$ (4 seconds), and temperature $\lambda = 0.1$. The encoder is trained using the Adam optimizer~\cite{kingma2014adam} for 25k iterations, with a batch size of 128 and a learning rate of $5 \times 10^{-5}$.
Our diffusion model uses 8 Transformer layers, a latent dimension of 512, and 1000 diffusion steps. Training is performed with a batch size of 128 on a single NVIDIA A100 GPU, completing within two days.

\noindent\textbf{Baselines.}
We systematically evaluate single and multi speaker motion generation models on the BEAT2~\cite{liu2024emage} dataset. CaMN~\cite{liu2022beat} introduced BEAT using a commercial motion capture system. DiffuseStyleGesture~\cite{yang2023diffusestylegesture} adopts a diffusion-based framework, while ZeroEGGS~\cite{ghorbani2023zeroeggs} uses a full-body style encoder for stylized motion. For fair comparison, we retrain and reproduce these methods under BEAT2’s standard. As RAG-Gesture’s~\cite{raggesture} code is unavailable, we report only its single-speaker results from the original paper.

\begin{table}[htbp]
\centering
\scriptsize  
\begin{tabular}{lccc}
\hline
Method & FGD~$\downarrow$ & BC~$\uparrow$ & Diversity~$\uparrow$ \\
\hline
GT & -- & 6.897 & 12.755 \\
CaMN~\cite{liu2022beat} & 0.804 & 6.704 & 10.935 \\
ZeroEGGS~\cite{ghorbani2023zeroeggs} & 0.730 & 6.890 & 10.768 \\
DiffuseStyleGesture~\cite{yang2023diffusestylegesture} & 0.829 & 7.380 & 11.783 \\
EMAGE~\cite{liu2024emage} & 0.666 & \cellcolor{mypale}7.861 & 12.742 \\
RAG-Gesture~\cite{raggesture} & 0.811 & 7.270 & 12.780 \\
GestureLSM~\cite{gesturelsm} & \cellcolor{mypink}0.408 & 7.140 & \cellcolor{mypale}13.240 \\
MimicParts (Ours) & \cellcolor{mypale}0.526 & \cellcolor{mypink}7.940 & \cellcolor{mypink}13.691 \\
\hline
\end{tabular}
\caption{\textbf{Quantitative comparison of the single-speaker model results on BEAT2.} Our method outperforms all baselines in BC and Diversity, achieving second-best performance on FGD. Red background indicates the best results, and yellow background indicates the second-best.}
\label{tab:comparison}
\end{table}

\begin{table*}
\centering
\scriptsize  
\setlength{\tabcolsep}{3pt}  
\renewcommand{\arraystretch}{1.2}  
\begin{tblr}{
  cells = {c},
  cell{1}{1} = {r=2}{},
  cell{1}{2} = {r=2}{},
  cell{1}{3} = {r=2}{},
  cell{1}{4} = {c=2}{},
  cell{1}{6} = {c=2}{},
  cell{1}{8} = {c=2}{},
  cell{1}{10} = {r=2}{},
  cell{1}{11} = {r=2}{},
  cell{3}{1} = {r=4}{},
  cell{6}{3} = {bg=mypink},
  cell{6}{8} = {bg=mypale},
  cell{7}{1} = {r=3}{},
  cell{7}{11} = {bg=mypale},
  cell{8}{4} = {bg=mypale},
  cell{8}{5} = {bg=mypale},
  cell{8}{6} = {bg=mypale},
  cell{8}{7} = {bg=mypale},
  cell{8}{9} = {bg=mypink},
  cell{8}{10} = {bg=mypale},
  cell{9}{3} = {bg=mypale},
  cell{9}{4} = {bg=mypink},
  cell{9}{5} = {bg=mypink},
  cell{9}{6} = {bg=mypink},
  cell{9}{7} = {bg=mypink},
  cell{9}{8} = {bg=mypink},
  cell{9}{9} = {bg=mypale},
  cell{9}{10} = {bg=mypink},
  cell{9}{11} = {bg=mypink},
  hline{1,3,7,10} = {-}{},
  hline{2} = {4-9}{},
}
           & Method\textbf{}  & FGD~$\downarrow$ & Upper           &                        & Lower           &                        & Hands           &                       & Diversity$\uparrow$ & SRA$\uparrow$    \\
           &                  &                  & $\text{BC}_u~\uparrow$ & $\text{Diversity}_u \uparrow$ & $\text{BC}_l~\uparrow$  & $\text{Diversity}_l \uparrow$ & $\text{BC}_h~\uparrow$ $\uparrow$ & $\text{Diversity}_l \uparrow$ &                     &                  \\
\shortstack{w/o Style\\Condition} ~ & GT               & –                & 4.838           & 0.803                  & 3.741           & 7.443                  & 5.848           & 0.563                 & 8.945               & -                \\
           & CaMN             & 0.667            & 3.296           & 0.628                  & 3.828           & 5.869                  & 5.152           & 0.365                 & 6.950               & -                \\
           & EMAGE            & 0.719            & 6.371           & 0.862                  & 3.360           & 7.765                  & 7.137           & 0.244                 & 8.467               & -                \\
           & GestureLSM       & 0.617   & 5.560           & 0.932                  & 4.408           & 8.593                  & 7.225           & 0.526                 & 10.209              & -                \\
\shortstack{w Style\\Condition}~   & ZeroEGGS         & 0.959            & 3.350           & 0.631                  & 3.782           & 5.834                  & 4.839           & 0.366~                & 7.010               & 54.77\%          \\
           & DiffuseStyleGesture & 0.970            & 6.500           & 1.054                  & 5.508           & 8.859                  & 7.164           & 0.695        & 10.825              & 51.99\%~         \\
           & MimicParts(Ours)       & 0.625            & 7.305  & 1.230         & 6.110  & 9.156         & 7.806  & 0.635                 & 11.302     & 72.54\% 
\end{tblr}
\caption{\textbf{Quantitative comparison of the multi-speaker model results on BEAT2.} We additionally report BC and Diversity for upper body (u), lower body (l), and hands (h). Our method outperforms all baselines in BC, Diversity, and SRA, ranks second in FGD, and demonstrates superior rhythmic consistency and motion diversity. Red background indicates the best results, and yellow background indicates the second-best.}
\label{tab:comparison_1}
\end{table*}

\begin{figure}[ht]
  \centering
  \includegraphics[width=0.6\textwidth]{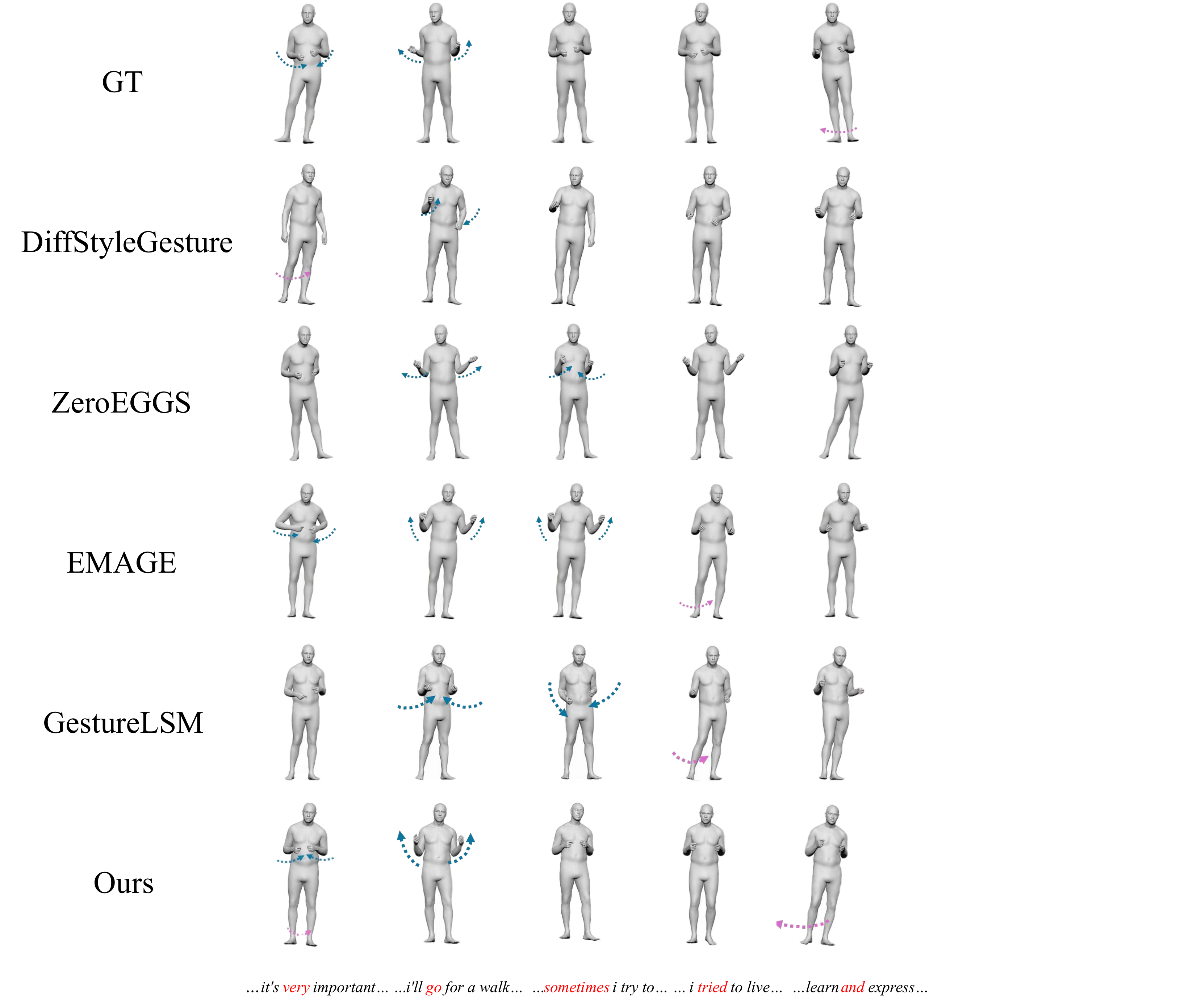}
  \caption{
    \textbf{Qualitative comparison on the BEAT2 dataset.} 
    The subject demonstrates limited lower-body movement and subtle upper-body gestures. 
    Our method generates highly synchronized full-body motion that closely matches the subject's natural style. 
    Blue arrows indicate the trend of hand movements, while red arrows represent the trend of leg movements.
  }
  \label{fig:3}
\end{figure}

\subsection{Quantitative Evaluation}
We adopt FGD~\cite{yoon2020speech} to evaluate the realism of the generated body gestures. To measure diversity, we follow~\cite{li2021audio2gestures} and compute the average L1 distance between multiple body gesture clips. Additionally, we use BC~\cite{li2021ai} to assess the synchrony between speech and motion. Inspired by the work of Jang et al.~\cite{jang2022motion}, we use a pre-trained style classifier to compute the Style Recognition Accuracy (SRA).

Stylized motion generation involves multiple speakers and styles, requiring multi-speaker training. For fair comparison, we retrain multi-speaker models on BEAT2. 

As shown in Table~\ref{tab:comparison}, under the single-speaker setting, our method significantly outperforms all baselines in terms of BC and Diversity, and achieves the second-best performance on FGD. These results strongly demonstrate the effectiveness of our approach in generating high-quality and diverse motion that is well synchronized with speech.

As shown in Table~\ref{tab:comparison_1}, in the multi-speaker setting, our method outperforms all baselines in terms of BC and Diversity. Although our FGD score is slightly higher than that of GestureLSM\cite{gesturelsm}, our method demonstrates superior rhythmic consistency and better preservation of motion diversity. To provide a more fine-grained analysis, we separately evaluate BC and Diversity for different body regions, reporting $\mathrm{BC}_u$, $\mathrm{BC}_l$, and $\mathrm{BC}_h$, where $\mathrm{BC}_u$ corresponds to the conventional BC metric. Similarly, we report $\mathrm{Diversity}_u$, $\mathrm{Diversity}_l$, and $\mathrm{Diversity}_h$ alongside overall Diversity. This detailed analysis highlights our method’s strength in achieving consistent coordination and motion diversity across all body parts.
Furthermore, our method significantly surpasses ZeroEGGS\cite{ghorbani2023zeroeggs} and DiffuseStyleGesture\cite{yang2023diffusestylegesture} in Style Recognition Accuracy (SRA), indicating a stronger capability in reproducing motion styles.

From a holistic perspective, our approach shows significant improvements in both style accuracy and motion realism, indicating its effectiveness in generating synchronized, coherent, and stylistically rich gestures.


\subsection{Qualitative Evaluation}
In this section, we conduct a qualitative comparison of two speech-driven stylized motion generation methods: ZeroEGGS\cite{ghorbani2023zeroeggs} and DiffuseStyleGesture\cite{yang2023diffusestylegesture}. 
We also compare the latest speech-driven motion methods such as EMAGE\cite{liu2024emage} and GestureLSM\cite{gesturelsm}.

Figure~\ref{fig:3} presents a qualitative comparison of generated gestures for subject ID=1, who exhibits a relatively static and conservative movement style characterized by small upper-body motions.
DiffStyleGesture~\cite{yang2023diffusestylegesture} captures stylistic cues to some degree but suffers from poor motion stability. ZeroEGGS~\cite{ghorbani2023zeroeggs} generates structurally stable gestures, yet its outputs often appear repetitive and lack expressive variation. EMAGE~\cite{liu2024emage} and GestureLSM~\cite{gesturelsm} align well with the speech rhythm, but their generated movements often lack personality and exhibit a low degree of stylization.

In contrast, our method generates gestures that are both temporally aligned with speech and faithfully reflect the subject’s conservative style. The motions are smooth, coherent, and spatially coordinated across body parts, demonstrating our model’s ability to preserve subtle style characteristics and maintain motion naturalness.

\begin{figure}[ht]
  \centering
   \includegraphics[width=0.5\textwidth]{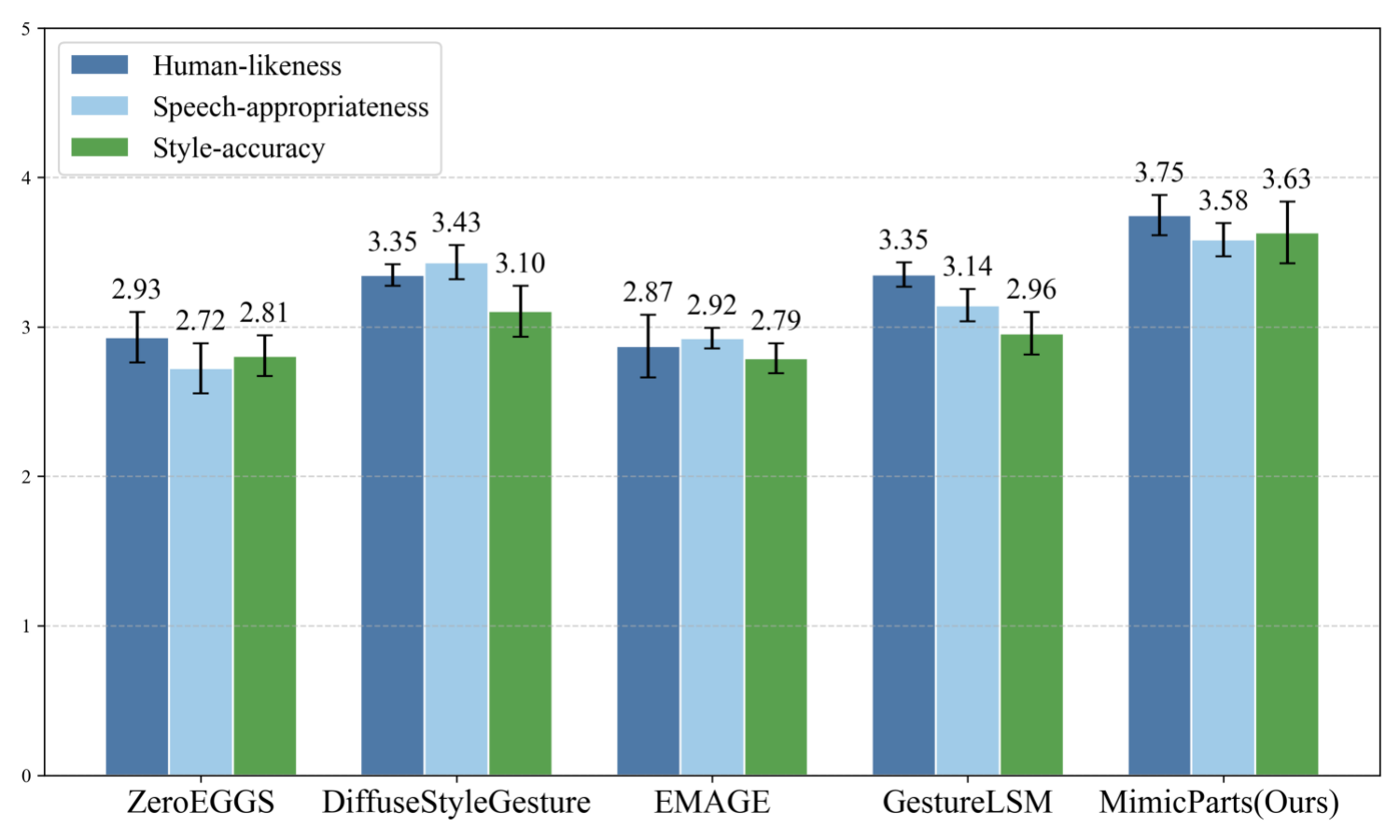}
  \caption{
\textbf{User Study.} Our MimicParts consistently receives higher user ratings across human-likeness, speech-gesture appropriateness, and style-accuracy, with a clear margin over all baselines.
  }
  \label{fig:5}
\end{figure}

\subsection{User Study}

We conducted a user study with 60 participants and 200 video samples -- 40 samples each from ZeroEGGS~\cite{ghorbani2023zeroeggs}, DiffStyleGesture~\cite{yang2023diffusestylegesture}, EMAGE~\cite{liu2024emage}, GestureLSM~\cite{gesturelsm}, and MimicParts -- to evaluate the perceptual quality of our results. Participants viewed the videos in randomized order and rated them based on three criteria: human-likeness, speech-gesture appropriateness, and style-accuracy.

For human-likeness, participants evaluated the degree to which the generated gestures resembled natural human motion in terms of realism and naturalness.  
For speech-gesture appropriateness, participants assessed whether the generated gestures were aligned with the accompanying speech in terms of rhythm and timing.  
For style-accuracy, participants assessed whether the generated motions were consistent with the given reference style.
As shown in Figure~\ref{fig:5}, MimicParts outperforms all baselines, generating motions that are more natural, speech-aligned, and stylistically consistent.

\subsection{Ablation Study}
To evaluate the effectiveness of our design, we perform ablation studies, with the results summarized in  Table~\ref{tab:ablation}.

\noindent\textbf{Effect of Part-aware Style Encoder.} Replacing the part-aware style encoder with a full-body encoder decreases performance in Diversity and SRA, as it lacks the ability to capture detailed differences across body regions, crucial for personalized motions.

\noindent\textbf{Effect of Rhythm and Emotion.} Disabling rhythm and emotion inputs lowers FGD and SRA performance, as the model’s ability to adapt motion style to speech rhythm and emotion is reduced. Incorporating these features enables more expressive and dynamic motions synced with speech rhythm and emotion.

\noindent\textbf{Effect of Rhythm.} Removing the rhythm feature's influence on motion style leads to an increased FGD distance and decreased BC, as the generated motion becomes less coherent and aligned with speech. 

\noindent\textbf{Effect of Emotion.} Removing the emotion feature's influence on motion style, the model struggles to generate expressive motions that align with the emotional state of speech, as seen in the decline of SRA.

\begin{table}[ht]
\centering
\scriptsize  
\caption{\textbf{Ablation Study.} Ablation study results comparing our method with various components removed. The table reports performance in terms of FGD, BC, Diversity, and SRA, with the best results highlighted in red.}
\label{tab:ablation}

\begin{tabular}{ccccc} 
\hline
                             & FGD~$\downarrow$                  & BC~$\uparrow$                     & Diversity~$\uparrow$               & SRA~$\uparrow$                       \\ 
\hline
Ours                         & {\cellcolor[rgb]{1,0.8,0.8}}0.625 & {\cellcolor[rgb]{1,0.8,0.8}}7.305 & {\cellcolor[rgb]{1,0.8,0.8}}11.302 & {\cellcolor[rgb]{1,0.8,0.8}}72.54\%  \\
w/o
Part-aware Style Encoder & 0.863                             & 3.296                             & 6.950                              & 58.86\%                              \\
w/o
Rhythm and Emotion       & 1.139                             & 6.575                             & 9.751                              & 45.60\%                              \\
w/o
Rhythm                   & 0.769                             & 6.762                             & 9.992                              & 60.38\%                              \\
w/o
Emotion                  & 0.880                             & 6.780                             & 8.624                              & 53.71\%                              \\
\hline

\end{tabular}
\end{table}

\section{CONCLUSION}
In this work, we propose MimicParts, a novel framework that integrates a part-aware style encoder with a part-aware motion latent diffusion model to generate high-quality, personalized 3D full-body motions. 

\noindent\textbf{Limitations.}
Current datasets assume static emotions per sequence and lack dynamic annotations, limiting the modeling of emotional transitions. Future work may explore large-scale datasets with rich emotion shift labels for more dynamic gesture synthesis.
\bigskip

\bibliography{aaai2026}

\end{document}